\documentclass{article}
\usepackage[utf8]{inputenc}
\usepackage[final]{neurips_data_2021}
\usepackage{hyperref}
\usepackage{xurl}
\usepackage{booktabs}
\usepackage{commands}
\usepackage{computed_values}
\usepackage{makecell}
\usepackage{graphicx}
\graphicspath{{figs/}}
\usepackage{placeins}
\usepackage{amsmath}
\DeclareMathOperator*{\argmax}{argmax} 

\title{ComSum: Commit Messages Summarization and Meaning Preservation}

\author{
  Leshem Choshen$^{*}$\\
  Department of Computer Science\\
  Hebrew University of Jerusalem\\
  {\tt\small leshem.choshen@mail.huji.ac.il}\\
  \And
  Idan Amit\thanks{First two authors contributed equally.}\\
  Department of Computer Science\\
  Hebrew University of Jerusalem and Acumen Labs \\
  {\tt\small idan.amit@mail.huji.ac.il}\\

}

\begin{document}

\maketitle

\begin{abstract}
We present ComSum, a data set of 7 million commit messages for text summarization. When documenting commits, software code changes, both a message and its summary are posted. We gather and filter those to curate developers' work summarization data set.
Along with its growing size, practicality and challenging language domain, the data set benefits from the living field of empirical software engineering.
As commits follow a typology, we propose to not only evaluate outputs by Rouge, but by their meaning preservation. 
\end{abstract}

\section{Introduction}

The amount of code written in the world is ever-growing.\footnote{Dataset in \url{https://figshare.com/articles/dataset/CumSum_data_
set/14711370}}\footnote{Code is found in \url{https://github.
com/evidencebp/comsum}}
When code is created by large groups of developers, documentation becomes essential and consequentially is a part of code quality \citep{santos2016judging}.
The need to communicate is especially important in distributed development, where shouting over the hallway cannot compensate for improper documentation.

Code development nowadays is usually supported by version control systems that track the source code modification, most commonly, Git. In Git,
each modification is called a \emph{commit}. A commit lists the changes in the code and a description by the developer. The description contains a one-line subject and a longer descriptive message.
Git and GitHub treat the subject line as a summary (See appendix \ref{ap:git_sum}), further incentivizing developers to do the same.
Hence, to build a text summarization data set, we use the subject as the summary and the rest of the commit message as the source.

In Section~\S\ref{sec:creation}, we describe the process of querying and filtering commits to curate ComSum, a commit summarization data set.
The data set is large, abstractive and  introduces a new domain for summarization (Dataset description in Section~\S\ref{sec:description}).
We consider several baseline results on the data set (see Section~\S\ref{sec:baselines}). The baselines include both neural summarization baselines and baselines based on relations between the subject and the message. Those shed light on the state of the art on the data set and the data set characteristics.
\doit{We are to first to use it so any result will be SOTA. I wrote later that we don't claim SOTA (since we didn't do multiple training). maybe initial results?}
Since commits are used to describe code changes, the taxonomy of changes is an important part of their meaning.
That enables us to evaluate a summary model by how well it preserves the meaning rather than by word overlap with a reference. We explain and provide an initial analysis in Section~\S\ref{sec:Meaning}.

\section{Related Work} \label{sec:related}
\begin{table}[]
\centering
\begin{tabular}{@{}lrrrr@{}}
\toprule
Dataset & \multicolumn{1}{l}{\# Docs.} & \multicolumn{1}{l}{Coverage} & \multicolumn{1}{l}{Density} & \multicolumn{1}{l}{Comp. Ratio} \\ \midrule
Arxiv/PubMed \citep{cohan2018discourse}& 346,187 & 0.87 & 3.94 & 31.17 \\
BigPatent \citep{sharma2019bigpatent} & 1,341,306 & 0.86 & 2.38 & 36.84 \\
CNN/DM \citep{Nallapati2016AbstractiveTS} & 311,971 & 0.85 & 3.47 & 14.89 \\
Newsroom \citep{grusky2018newsroom} & 1,212,739 & 0.83 & 9.51 & 43.64 \\
XSum \citep{narayan2018dont} & 226,677 & 0.66 & 1.09 & 19.25 \\
BOOKSUM Paragraph \citep{kryscinski2021booksum} & 142,753 & 0.5 & 0.92 & 6.47 \\
\hline
ComSum & 7,540,026 & 0.27 & 0.89 & 8.1 \\ \bottomrule
\end{tabular}
\caption{ComSum is more abstractive, as seen in low coverage and density.\label{tab:related}}
\end{table}

Textual Summarization has become a central task in Natural Language Processing, attracting pretrained models \citep{zhang2020pegasus, qi2020prophetnet} and specialized models \citep{dou2021gsum}.
With this came also a need for larger, more diverse and more challenging data sets.

News is the most common domain for summarization data sets \citep{over2007duc, Nallapati2016AbstractiveTS}. Some news data sets \citep{grusky2018newsroom} emphasize size and abstractiveness over extractive summarization, not favoring copying the source \citep{narayan2018dont}. While news is advocated for its general domain, we find the vocabulary, which should demonstrate it, is rather low. The vocabulary of the commits is over 2M in the validation set alone (to be fair in terms of size) and 19M overall (top reported is 1.4M NYTimes dataset \citep{narayan2018dont,sandhaus2008new}). Similarly, the vocabulary of the summaries is 0.5M and 3.9M (NYTimes 0.3M).

\citet{kryscinski2021booksum} also called for more challenging and diverse abstractive data sets. We compare (Table~\ref{tab:related}) the abstractness as measured by low density and coverage \citep{grusky2018newsroom} finding our dataset excels in that regard.
Others concurred, offering large datasets of different domains 4M crawled TL;DR from Reddit \citep{volske2017tl} and 1.3M patents \citep{sharma2019bigpatent}.

Our work follows all those desirable traits. It is more abstractive, it introduces a new natural summarization domain, it is even larger than current data sets and will keep growing in size substantially.

In addition to the data set, we propose (see Section~\S\ref{sec:Meaning}) an evaluation procedure specific to the domain at hand that emphasizes the overall meaning of the summary rather than its similarity to a reference as commonly assessed by summarization evaluation.

Apart from summarization, this work also contributes to the growing field of NLP for aiding programming. This emerging field includes tasks such as code generation \citep{hayati2018retrieval}, search \citep{gu2018deep}, automatic documentation \citep{miceli2017parallel} and updating documentation by code changes \citep{panthaplackel2020learning} and also has a dedicated workshop \citep{nlp4prog}.

Evaluation of summarization mainly focuses on general quality of summarizations \citep{bhandari2020evaluating, zhang2019bertscore}, with some exceptions \citep{Wilber2021ToPO, Xu2021DissectingGM}. Some work showed hallucinations are a problem \citep{kryscinski2020evaluating} and focused on evaluation of factual consistency \citep{gabriel2020go, honovich2021q, pagnoni2021understanding}. Other fields of text generation provide additional ways to extract informative measures \citep{ribeiro2020beyond}. Measures that tell about certain characteristics of the output, rather than bottom-line scores. Such methods include evaluation on minimal changes to the input \citep{warstadt2020blimp}, challenge sets \citep{macketanz2018fine, choshen2019automatically}, metrics dedicated to specific characteristics such as grammaticality \citep{vadlapudi2010automated} or meaning preservation \citep{choshen2018reference}, manual evaluation \citep{graham2015accurate}, evaluation dependent on the domain data \citep{Choshen2018AutomaticMV}, understanding the inner workings of networks \citep{tenney2019bert, slobodkin2021mediators, voita2020analyzing} and more.

Several data sets and tasks share similarities with summarization. Those include simplification \citep{alva2020asset}, sentence compression \citep{filippova2013overcoming}, web-page snippet generation by query \citep{chen:2020a} and single sentence summarization \citep{rush2015neural}.

\section{Data Set Creation}\label{sec:creation}
In this section, we describe the creation of the proposed data set, ComSum. Specifically, we describe how we acquire and filter projects and commits to extract reliable summarizations.

Open source code is shared in large amounts on different platforms.
GitHub, owned by Microsoft, is a large hosting service for projects using the Git version control system.
In 2018 GitHub announced that they hosted 100 million projects \citep{warner100mrepo}.%

We base our data set on the BigQuery GitHub schema \citep{bigQuery}. The schema allows querying commits pushed to GitHub by various metadata attributes.
The BigQuery GitHub schema contains about 3.4 million \emph{public} projects prior to 2021,
but the vast majority are not appropriate for studies of software engineering, being small, non-recent, or not even code.

\subsection{Projects selection}
Software is written in projects or repositories, sharing a general goal. While some projects consist of a few files shared, others are constantly updating. We focus on the latter, aiming to filter the first to avoid commits where clear communication is neither crucial nor enforced.

Therefore, the main effort in the data set construction is to identify projects of interest and filter irrelevant ones.
We base our filtering on the methodology developed by \citet{Amit2021CCP} to study software code quality.
First, we choose only large enough and up to date projects by requiring at least 50 commits during 2020\footnote{ Projects requiring more or updated data should keep the requirement of 50 commits per year. Instead, they can choose a set of years instead of just 2020, and use projects that were active in at least one of those. They can also reduce other filters (e.g., the 100 characters difference).}.\idan{I find this footnote confusing. Can we remove it?} \leshem{It is quite possible that if people will actually use it, someone in the future would want to recreate, and this would be important. I rephrased it, is it better?}
Note that this is less than one commit per week, a rather low bound filtering tiny projects.
However, this step alone was enough to reduce the number of relevant projects to 32,562, 0.96\% of the prior step.

The next step is the removal of redundant projects.
Github enables `forking': copying a project, sometimes for investigation and sometimes for modifying without altering the main project.
We identified forks using the GitHub API and removed them from our data set.
We also removed projects that shared too many commits with a more popular project, having more stars, in order to use `Spark' of the `Apache Software Foundation' and not a hobbyist project built upon it.
This step ended with 25,535 projects, 78\% of the prior step.

We filtered projects not likely to be software projects, identified by the lack of bugs. A project without commits fixing bugs was identified by a negative Corrective Commit Probability (CCP) \citep{Amit2021CCP}.
Projects with negative CCP are less likely to be software projects, externally validated by programming languages identification and imply no use of the pull request development mechanism.
After filtering projects with invalid CCP we were left with 22,820 projects, 89\% of the previous step.
Out of these, \projectsNum projects had commits fitting the conditions described next, 86\% of the previous step and 0.57\% of overall projects.

\subsection{Commits Selection}\label{sec:selection}

When constructing a data set, the intended use is of high importance.
In our case, many future uses are possible.
One might want to remove single-person projects since they do not represent communication.
Others might be interested only in single-person projects since they represent documentation for a future self.
We choose not to add needless filtering here in order to allow future flexibility.

We used only two constraints.
We require that the message will be at least 100 characters longer than the subject.
While the value of 100 is arbitrary, a significant gap in the length is required in order to have a meaningful summarization.

We considered only commits earlier than 2021 to help future reproducibility.
Note that this condition is not enough to guarantee the stability of the data set since existing projects might be deleted and their content will be removed from the index.
In order to cope with that, we release a static version of the data set together with the code to extract it.

Overall, the data set has \mainDataSetSize commits.
A commit might appear in a few repositories, having the same subject, message and content.
Since we are interested in text summarization, those are considered duplicates and we remove those 16\% of the data that was repetitive. This step is especially important to prevent training data from leaking to the test set.

A message or a subject might appear more than once.
For example, the most common subject is ``Updating submodules'', appearing 6612 times.
However, 96\% of the subjects are unique.
Also, a message and its summary never repeat.
We left the recurring message since this represents software development, researchers looking for uniqueness can remove those.
We provide appearance distribution and common messages in our released code repository (App.~\S\ref{sec:sup}).

0.8\% of the subjects appeared in the rest of the message.
Manually analysing the common ones suggest those are generic subjects.
The leading ones are 'WebCore:' (1554 times), `Updated Spanish translation.' (809 times) and `Updated Norwegian bokmål translation.' (347 times).
Again, to ensure maximal future flexibility we left them.

Another question of interest is to what extent, the subject represents a summary of the message.
The user interface aim to such usage but the developers are free to choose the any content.
Both authors independently manually labeled 100 samples (See labeling protocol App.~\S\ref{ap:guidelines}).
The labeling agreement was 82\%, and after discussion and tuning of the protocol it reached 99\%.
In essence, we require a proper summary to add details or reason \doit{changed from "add no new information"} and to contain the gist.
We also labeled whether the summary was generic or specific to the commit.
80\% of the subjects were a proper summary. Out of the rest: 35\% had administrative message (e.g., just the developer detail).
20\% had a subject that indicates a merge (e.g., merge branch 3.4) and the message as the content.
In 15\% the subject was by reference (e.g., fixed \#123)
and in 5\% subject was generic.
The rest of the 25\% are diverse and harder to identify and handle. We next discuss filtering those cases.

We provide a list of 429K merge commits identified by Git as having more than one commit parent.
Note that some do have an informative subject.
We also provide a \emph{heuristic} for identifying administrative messages.
We identify them by the distance of administrative terms (e.g., 'Signed-off-by:', 'Change-Id:', the full list is found in the supplementary materials)
from the beginning of the message. Manual labeling (See App.~\S\ref{ap:ad_heur}) shows those to have 98.9\% precision and 75\% recall.
Filtering both cases leads to about 90\% of the subjects serving as summaries on our labeled sample.
Nonetheless, we provide the entire data set, to maximize future uses and provide code for filtering.
\idan{Removed the next since it redundant}
To reassess, we labeled 200 unfiltered commits, only 26 (87\%) were not strictly summaries.
The performance results described later are based on the filtered data set.

\section{Data Set Description}\label{sec:description}

Following these procedures, we create ComSum, a data set of commit messages and their summarization.
ComSum contains \mainDataSetSize commits from \projectsNum projects, written by \authorsNum authors.

In addition to the summarization objective, we publish metadata on each commit.
Each record also has the commit hash, a unique identifier that can be used for future work.
Last, we provide the name of the repository from which the commit was taken. We note that a commit might be included in several repositories (projects), but we choose only one (see Section~\S\ref{sec:creation}).

Our guideline when constructing ComSum was to ease consistent use.
Hence, each record contains the subject, the message without the subject and the entire commit message combining the two.
While it almost doubles the storage, it prevents future noise due to different implementations.

The average subject has \avgSubject characters, the average message has \avgMessage characters.
The average compression rate, the commit message length divided by its subject length is \avgCompress.

We separate ComSum into train, validation and test to compare future results consistently.
We provide a main split by the repository and a minor split by the commit.
The separation is based on hash functions so it is both pseudo-random and reproducible.
The test set and the validation set have about 472K examples each.

Note in table \ref{tab:baselines} that there are differences between the performance on the test and train sets.
In general, we observe that the test set is slightly harder.
Note both \textit{Zero-Shot Bart} and \textit{Random Message Sentence} have RougeL on the test set that is lower in about 3 points than on the train set.
Since both of them do not require any training, it is due to the difference between the data sets. As the goal in this split by \emph{the repositories} was to prevent data leakage.
Hence, the repositories in each splits are of different domains, community size, etc. and therefore some dissimilarity is acceptable.
Indeed, comparing their metrics we note test is more abstractive with a coverage of 0.25 for test compared to 0.31 and test density of 0.86 compared to 0.99.

Overall, 97\% (89\% without the optional filtering) of test set subject lines never appear in the training set. Other lines are common and appear repeatedly in the train set (e.g., Merge branch 2.7 into 2.8). However, manual inspection suggests their corresponding commit messages share little resemblance. As the model should learn when to recall more generic summaries as well as to generate specific ones well, we leave those in the test set and do not separate train and test by subjects.
We also create subsets of meaning-preserving commits, explained in Section~\S\ref{sec:Meaning}.

\section{Baselines}\label{sec:baselines}

We inspect the behavior of neural models and baselines. Those provide insight on the characteristics of the data set, set a baseline for future work and allow for unique evaluation motivated by domain knowledge (see Section~\S\ref{sec:Meaning}).
In all experiments, we compute Rouge 1,2 and L scores \citep{lin2004rouge}.

For a neural model, we used BART \citep{lewis2020bart}. We consider two variations, untrained for zero-shot performance and fine-tuned on the train data set (See hyperparameters in App.~\S\ref{ap:hyperparams}).
BART was originally trained in the domains of Wikipedia and Books and it was not exposed to the non-formal and technical language found in commit messages.
However, BART, pre-trained that way, showed impressive results even on domains such as malware detection \citep{oak2019malware}.

Regardless, BART results are high (Table~\ref{tab:baselines}) and it surpasses Zero-shot results by a large margin, suggesting the data set is large enough to overcome the domain shift at least partially.
BART \citep{lewis2020bart} achieved RougeL of 44.2 on the CNN and Daily Mail data sets \citep{48562} and 27.2 on the XSum data set \citep{narayan2018dont}, suggesting ComSum is still challenging.

Following previous work \citep{kryscinski2021booksum}, we provide heuristic summarization techniques for analysis.
Results were computed on 10k samples and presented in Table \ref{tab:baselines}.
These heuristic summarizing techniques do not learn and therefore the train and test splits are irrelevant to them.

As a first, \textit{Subject and Message} do not summarize at all and include both the subject and the message, acquiring a \subjectvsMessageRougeL RougeL.
This demonstrates the need for compression.
Note that this method cannot be used for prediction since it uses the summary.

\begin {table}[h!]\centering
\begin{tabular}{ | l| l| l | l |l |}
\hline
Model & Data set & RougeL & Rouge1 & Rouge2\\
\hline
Bart & Train & 33.2 & 35.3 & 19.2\\ \hline
Subject and Message & Train & 29.5 & 29.5 & 26.7 \\ \hline
Bart & Test & 27.2 & 29.0 & 15.4\\ \hline
Zero-Shot Bart & Train & 17.6 & 19.7 & 7.8\\ \hline
Related Fix & Train & 15.5 & 15.6 & 11.9 \\ \hline
Zero-Shot Bart & Test & 14.9 & 16.8 & 6.5\\ \hline
Related Commit & Train & 14.6 & 14.9 & 10.4 \\ \hline
Random Message Sentence & Train & 13.6 & 15.3 & 6.4\\ \hline
Message without Subject & Train & 12.0 & 13.9 & 6.1 \\ \hline
Random Message Sentence & Test & 10.1 & 11.4 & 4.5\\ \hline
\end{tabular}
\caption{ \label{tab:baselines} Baselines results on different data sets. Training on the data set provides a significant boost. Repeating the commit message or a related subject is not enough for a high score.}
\end{table}
\doit{Won't it be better to sort the table by RougeL? Do you want to group models?}\leshem{Group models, definitely, sort between them by rougeL, good idea! (changed, do you think it is better to split the train-test BART (put subject and message in between as it fits there by score?)}
\idan{I changed to by score only, otherwise it is confusing. If you prefer grouping, than all train/test pairs should be together}
We perform another such test using the \textit{Message without Subject} as the summarization. This method reached a RougeL score of \nosubjectvsMessageRougeL which is relatively low. This implies that while there is information in the commit message, repeating it is far from enough to achieve a good summarization score in ComSum.

Similarly, we define a \textit{Random Message Sentence} baseline.
We split the message into sentences and randomly pick a single non-empty sentence as the summarization, which achieves a RougeL of 13.6.
This comes to see how well a more reasonably sized extraction of the input would do (on average, messages are \avgCompress times longer than subject).
As expected, it is worse than the whole message and shows sentence extraction is unlikely to be considered a good commit summarization.

Another test case is \textit{Related Commit}. We extract pairs of commits by the same author, in the same project, within a week's range of each other.
We consider the subject of one commit as the summary of its pair. Simulating summaries of the same person on the same topic, regardless of the specific message. We expect high scores from such a measure if summaries are general and quite similar or even repetitive upon related commits.
The related commit subject benchmark reached a score of \relatedCommitRougeL, suggesting this is not the case. Where we require both commits to be a bug fix, a setting we term \textit{Related Fix} the results are higher. Results are also somewhat higher than those achieved by extracting a \textit{Random Message Sentence} from the commit.
This shows that the topic and style conserve some of the meaning needed for the summary, but they are far from satisfactory surrogates of a summary.
Please note that in the text summarizing we treat each commit on its own and compare the commit message and subject.

Calculating the measures on a random sample, we confirm the difference stems from the split of different repositories to train and test data.
Bart, which is a trained model drops in 6 points on the test split by \emph{repositories}.
It is common to have over-fitting due to memorization \cite{feldman2020does,arpit2017closer}.
However, when we compared the results of BART on the spilt by \emph{commits} the drop was of only one point, so overfit is at most small.

Surprisingly, BART achieves similar results to that of the message and subject, which is a summarization that includes all needed summary (the actual reference) and a lot of unneeded text.

Manually inspecting the outputs of BART shows mixed results. On the one hand, a reasonable percentage of sentences resemble the reference and in general convey the right action done in the commit. On the other hand, many errors are found. Even well-structured sentences fail in terms of factual consistency. The network hallucinates terms, names and numbers. For example, "Merge pull request \textbf{\#14}" instead of \#1110, "Bump flake8-isort from 2.9.0 to \textbf{2.8.1}" instead of to 2.9.1 and other more complex cases.
The high Rouge score co-existing with common mistakes calls for evaluation procedures to differentiate allowed sentence variants from outputs of wrong meaning.

\section{Meaning Preserving Summarization}
\label{sec:Meaning}

Meaning preserving is part of the definition of the text summarization problem \cite{gupta2010survey,chopra2016abstractive}.
\citet{gupta2010survey} suggested an elegant mathematical definition $$model\ summary = \argmax_x p(message|x)$$

While attractive, this definition suffers from several drawbacks.
The choice of the probability model is subjective.
The computation of the most likely summary might not be feasible.
Last, it is not clear if the writer intends and is capable of summarising the message that way, meaning that the samples that we use do not fit the concept aimed to learn.

Testing summarization quality by word overlap alone might be unreliable \citep{choshen2018inherent} and human annotation is costly.
Fine-grained factual consistency is less specific to this domain and is an active field of study \citep{gabriel2020go, honovich2021q}.
We hence, provide specialized test approaches, asserting that the output summaries preserve the meaning.

There are many aspects of meaning that one might choose to preserve.
For example, the sentiment of the author, the register of the language, etc.
A good property to preserve should satisfy  few conditions.

First, there should be an agreement that this meaning should be preserved.
Given a model that does not preserve sentiment, one might claim that this is desirable, leading to a more concise summary removing irrelevant information.

The second property should be that the aspect can be estimated using a computable function, a requirement for automation on a large scale.
The aspect should be as objective as possible, (e.g., as measured by agreement between human annotators), in order to avoid a model that has a different subjective ``point of view''.

Our data set enables the use of the commit type, having these properties.
We use the classical commit taxonomy of Swanson, suggested in 1976, classifying commits as: corrective (aka bug fix), adaptive (adding new features) and perfective (refactoring and documentation improvements) \citep{swanson1976dimensions}.
This taxonomy is very common among developers and software engineering researchers \citep{mockus2000identifying}.
Therefore we used a model for corrective, adaptive and refactor, a subset of perfective. We chose to focus on the latter as refactor changes are related to the source code and are therefore more important to communicate.
The classification captures the essence of the work done in the commit, hence, its meaning should be preserved.

A commit might be tangled and serve several goals, for example, both fix a bug and refactor the fixed code \citep{6624018, herbold2020largescale}.
Other than being less common, being a bug does not influence being a refactor and both meanings should be preserved.

Human annotators reach an agreement of 95\% on the classification of a commit as a bug \citep{Amit2021CCP}.
We use the classifiers of \citep{10.1145/3345629.3345631, Amit2021CCP}, reaching accuracy of 93\% for corrective and refactoring
, very close to the human level, and the adaptive classifier of accuracy 65\%.
Hence we are capable of estimating the classification at scale accurately.

\looseness=-1
One could use the classification of commits as the meaning to preserve.
However, a naive model identifying a list of core terms
reaches an accuracy of 88\% classifying the corrective concept.
Since these words are common in commit messages, a model ignorant of the meaning might still use them as a summary. Therefore, we suggest more challenging cases for meaning preservations analysis.
We compute for all the discussed concepts the precision-like meaning-preservation metric, $P(concept(model(message)))|P(concept(message))$.
BART's highest precision for any of the concepts we tested on was 75\%.
This emphasizes how common and severe non-preserving summaries are and it calls for further investigation.
However, an alternative claim is that omitting the concept from the summarization is fine since it is not important.

In order to cope with this claim, we construct cases in which the classification as the concept is added and not omitted.
We use the term `core term' for terms whose appearance is indicative of the concept, like 'bug', 'bugfix', 'error', 'fail', and 'fix'.
The lift of being classified as corrective given the appearance of a corrective core term is 99\%, for adaptive the lift is 60\%, and for refactor 238\%.
A naive model will fail on sentences having a core term yet not classified as the concept.
Examples are ``Added \emph{error} handling'', ``Used \emph{fixed} point arithmetic'', ``This is not a \emph{bug fix} but a new requirement'', etc.
We built a suitable data set, we selecting messages that contain a core term, yet are still classified as negative by the concept's classifier. I.e., they contain a core term that usually suggests they belong to one concept, but they do not.

Hence, in order to preserve the meaning, the summary should match the message in concept.
In that case, the system output should be of the negative class too and preferably contain the core term.

Before evaluating meaning preservation, we present the Rouge score on the meaning preserving data sets.
Comparing the results in Table \ref{tab:baselines} and Table \ref{tab:meaning-preserving-benchmark}, shows similar performance trends.

\begin {table}[h!]\centering

\begin{tabular}{ | l| l| l | l |l |}
\hline
Model & Data set & RougeL & Rouge1 & Rouge2\\
\hline
Bart & Adaptive & 26.6 & 28.6 & 14.5\\ \hline
Bart & Refactor & 26.9 & 28.8 & 14.1\\ \hline
Bart & Corrective & 26.2 & 28.3 & 13.6\\ \hline
Zero-Shot Bart & Adaptive & 14.3 & 16.3 & 5.7 \\ \hline
Zero-Shot Bart & Refactor & 16.0 & 17.7 & 6.5 \\ \hline
Zero-Shot Bart & Corrective & 15.9 & 18.1 & 6.8\\ \hline
Random Message Sentence & Adaptive & 9.9 & 11.0 & 4.2\\ \hline
Random Message Sentence & Refactor & 10.1 & 11.5 & 4.4\\ \hline
Random Message Sentence & Corrective & 9.6 & 10.9 & 3.5\\ \hline
\end{tabular}
\caption{ Rouge scores on typed test sets. Trends are similar to those on the general test set. \label{tab:meaning-preserving-benchmark} }
\end{table}

\begin {table}[h!]\centering
\begin{tabular} { | l| l| p{15mm}| p{15mm}| p{15mm}| p{15mm}| p{15mm}  | }
\hline
\textbf{Model} & \textbf{Data set} & \textbf{Not Preserved} & \textbf{Core and Concept} & \textbf{Not Core and Concept} & \textbf{Core and Not Concept} & \textbf{Not Core and Not Concept}\\
\hline
Bart & Corrective & \textbf{0.10} & 0.07 & 0.03 & 0.10 & 0.79\\ \hline
Bart & Refactor & \textbf{0.16} & 0.10 & 0.06 & 0.09 & 0.75\\ \hline
Bart & Adaptive & \textbf{0.35} & 0.30 & 0.05 & 0.13 & 0.51\\ \hline
\end{tabular}
\caption{ \label{tab:semmantic-corrective} Meaning Preserving on summaries containing a distractor core term (e.g., "bug") not fitting their concept type (e.g., corrective). Models should preserve meaning and though the distraction. Indeed, those cases are confusing for the model. }
\end{table}

However, the meaning-preserving property allows us to extend our analysis beyond this bottom line.
Table \ref{tab:semmantic-corrective} presents the classification of the summaries of the meaning-preserving messages that have a core term of a concept yet are not part of the concept's class.
The message  ``Added \emph{error} handling'' is not classified as a bug fix despite the appearance of the core term ``error''.
When a message contains a core term but is still classified as having a concept, it indicates that the concept is indeed not the core's one, as the prior on matching concept and core term is very high.
We build such data sets for corrective, refactor and adaptive concepts in order to demonstrate it is not a property of a specific concept.

Next, we observe the summaries generated by Bart.
When the summaries include a core term, yet are not classified as discussing a concept, the meaning is correct, matching the message.
This is the best case where the summary matches both the concept and the core term. Optimally, all summaries would fall under this case.

When there is no core term and the summary is not classified as a (wrong) concept, it might be a good summary in terms of meaning, not stating the nonexistent concept.
On the other hand, these cases might be a result of hallucinations, as they do not mention the core term.

However, when there is a core term and the summary is classified as the concept, then the meaning was changed.
Last, when there is no core term and the message is classified as discussing the concept, not only the meaning is changed, the core term disappears and the summary might be a result of hallucinations too.
\textit{Not Preserved} in the table represents the cases where the meaning was changed. It is the sum of \textit{Core and Concept} and \textit{Not Core and Concept}. We find that 10-35\% of sentences checked, change their meaning.
These meaning-preserving probabilities serve as quality metrics for the summary. Future work may integrate them into the loss function, forcing the model to both produce the right words and keep the meaning.

It is worth reiterating that the commit classifiers are not perfect.
While they were trained on thousands of messages, the meaning preserving messages are different in nature and the distracting core term makes them harder to classify.
We manually labeled 20 messages for each concept, in order to estimate how many of the messages in the sub data sets indeed have the desired properties.
For the adaptive labels, all had the core term and 60\% were not adaptive, fitting for meaning preserving.
For refactoring, only 35\% of the labels fit the meaning preserving data set, and in the corrective 75\%.
We use the same classifiers for the message and the summary, mitigating the accuracy importance.
However, when comparing results between data sets, the classifier accuracy should be taken into account.
Assuming independence, estimating mistakes with $P(Not\;Preserving)*(1-Accuracy)$, which is much higher in corrective and adaptive compared to refactor.

The fact that meaning is often not preserved is likely to be general.
We used off-the-shelf pre-trained models.
Moreover, we did not train the model directly to preserve the meaning, rather we trained it to generate the summary token by token.
Thus, the models are trained in a rather general way one that is not specific to summarizing commits or to preserving meaning. The only change is that the models were trained on data which meaning could be evaluated on. Thus, we rely on the distinctions known in software engineering research to evaluate the behavior of current models expecting it to be relevant to other summarization domains as well.


\section{Limitations and Threats to Validity}\label{sec:limitations}

We discuss in this section the limitations of our works. We also discuss the ethical concerns at App.~\S\ref{sec:ethical}

The data set is based on active open source projects.
These projects will keep advancing and in the future will have more commits that will enable building larger data sets.
We extract the current commits and freeze them to enable reproducibility.
We also provide the extraction code and limit the commits to commits earlier than 2021.
However, the current data set might not represent future development activity.
A data set that will be generated in the future might not match the frozen data sets since projects that will be deleted will not be included.

For the meaning preserving analysis, we use commit classification models.
Each model has different biases and prediction performance.
We believe that further improving current models and using more models will reduce the influence of current models weaknesses.
In turn, this makes the model exact details part of the reporting and reproducibility.

As an external validity concern, it is not clear how much the projects in the data set represent open source development.
While we control our project selection, we could not find documentation explaining how projects are selected into the BigQuery schema that we rely on.
Some well-known projects are not included (e.g. The Apache Software Foundation's Mahout and ActiveMQ).
An absence that is even harder to explain is that of Microsoft's VSCode, an extremely popular editor with more than 100K stars.
It existed and was later removed, though the project is still publicly developed.
On the other hand, our data set contains \projectsNum projects, more than the usual amounts used in research on the GitHub schema: 7,557 \cite{Amit2021CCP}, 1,531 \cite{10.1145/3345629.3345631}, and 677 \cite{amit2021follow}.

Git enables developers to create commits in a dedicated `branch' and then `squash' them into a  single `merge' commit.
The default message of the squashed commit is the concatenated messages of all the branch commits.
While all the commits in a branch are related, the cohesion is lower than in a regular commit and the messages are longer.
These cases can be identified, either by filtering merge commits or simply long messages.
We want to allow future researchers maximal flexibility and therefore we just alert on this issue instead of enforcing a specific filter.

Our data set is specific to software development.
Hence, improvement on Comsum might not generalize to other domains.
Clearly, bugs and refactoring will not appear in other domains.
Other than this obvious difference, a high percentage of software developers are males \cite{terrell2017gender}, raising another external validity threat.

Our choice to set a minimal difference of 100 characters between subject and message is not the only option.
48 million commits, 94\% of all of the commits in our projects, have a message longer than their corresponding subject.
Our choice led to an average ratio of $\frac{len(message)}{len(subject)}$ = \avgCompress, requiring significant compression.
In this case, we did not provide the messages with a smaller difference since that will require much higher storage of less interesting or even misleading messages.
The code that we provide enables others to generate similar data sets to their taste.

Training models is costly.
Therefore we could not repeat the training on many samples in order to provide error bars for the benchmarks.
However, we evaluate the performance on large test sets.
Table \ref{tab:baselines} shows that both Bart, Zero-Shot Bart and `Random Message Sentence' get very close results on the train and test which is another reason to believe results are robust.

\section{Future Work}
\label{sec:future}

The commit data set has the important property of task type meaning-preserving.
This property enables requiring and evaluating beyond lexical similarity.
It will be interesting to identify such properties in general texts.
For example, forbidding a change from a positive sentiment to a negative one (e.g., in dropping the `not' in `not bad') might be a general property.
Negation, modals, and idioms seem to be a suitable area to find such properties.
Text topics, like security or performance in commit messages, might be suitable for meaning preserving too.

Our data set is also useful as a testbed for active learning.
In active learning, there is a large amount of unlabeled data and the goal is to find a small group of informative samples that can be labeled in a feasible cost \cite{Settles10activelearning}.
One can use labeling functions \citep{NIPS2016_6523, archimedes}, computational functions that are weak learners \citep{schapire1990strength}.
For example, commits not classified as corrective, perfective or adaptive, are assured to be a false negative of one of the classifiers.
The method was used to boost the corrective commit classifier model \citep{Amit2021CCP}.

One can use the \projectsNum projects for topic detection, a data set that is expected to be challenging since all the projects deal with software and hence the topic difference is more delicate.
Another possible use is to enhance the data set with author identifier, and use pairing \citep{amit2019machine} in order to learn author writing style.

\section{Conclusions}
\label{sec:Conclusions}

We present a text summarization data set, ComSum, of  significant size, and a methodology to extract larger such data sets in the future. ComSum is not only of a large size, it provides new challenges such as summarizing in a new domain, where a lot of terms appear and constantly change.

We present benchmarks based on related messages, allowing us to assign meaning to a model performance evaluation.
We also identify meaning-preserving properties that enable training and evaluating models on goals beyond lexical similarity.

\bibliographystyle{plainnat}
\bibliography{bibtex.bib}

\begin{thebibliography}{70}
\providecommand{\natexlab}[1]{#1}
\providecommand{\url}[1]{\texttt{#1}}
\expandafter\ifx\csname urlstyle\endcsname\relax
  \providecommand{\doi}[1]{doi: #1}\else
  \providecommand{\doi}{doi: \begingroup \urlstyle{rm}\Url}\fi

\bibitem[Alva-Manchego et~al.(2020)Alva-Manchego, Martin, Bordes, Scarton,
  Sagot, and Specia]{alva2020asset}
Fernando Alva-Manchego, Louis Martin, Antoine Bordes, Carolina Scarton,
  Beno{\^\i}t Sagot, and Lucia Specia.
\newblock {ASSET}: {A} dataset for tuning and evaluation of sentence
  simplification models with multiple rewriting transformations.
\newblock In \emph{Proceedings of the 58th Annual Meeting of the Association
  for Computational Linguistics}, pages 4668--4679, Online, July 2020.
  Association for Computational Linguistics.
\newblock \doi{10.18653/v1/2020.acl-main.424}.
\newblock URL \url{https://www.aclweb.org/anthology/2020.acl-main.424}.

\bibitem[Amit(2021{\natexlab{a}})]{Amit2021Analysis}
Idan Amit.
\newblock Analysis utilities.
\newblock Aug 2021{\natexlab{a}}.
\newblock \doi{10.5281/zenodo.5179861}.

\bibitem[Amit(2021{\natexlab{b}})]{Amit2021CommitClassification}
Idan Amit.
\newblock Natural language processing for software engineering.
\newblock Aug 2021{\natexlab{b}}.
\newblock \doi{10.5281/zenodo.5179812}.

\bibitem[Amit(2021{\natexlab{c}})]{Amit2021General}
Idan Amit.
\newblock General infrastructure for github data analysis over bigquery.
\newblock Aug 2021{\natexlab{c}}.
\newblock \doi{10.5281/zenodo.5179783}.

\bibitem[Amit and Feitelson(2019)]{10.1145/3345629.3345631}
Idan Amit and Dror~G. Feitelson.
\newblock Which refactoring reduces bug rate?
\newblock In \emph{Proceedings of the Fifteenth International Conference on
  Predictive Models and Data Analytics in Software Engineering}, PROMISE'19,
  page 12–15, New York, NY, USA, 2019. Association for Computing Machinery.
\newblock ISBN 9781450372336.
\newblock \doi{10.1145/3345629.3345631}.
\newblock URL \url{https://doi.org/10.1145/3345629.3345631}.

\bibitem[Amit and Feitelson(2021)]{Amit2021CCP}
Idan Amit and Dror~G. Feitelson.
\newblock Corrective commit probability: a measure of the effort invested in
  bug fixing.
\newblock \emph{Software Quality Journal}, pages 1--45, Aug 2021.
\newblock ISSN 1573-1367.
\newblock \doi{10.1007/s11219-021-09564-z}.
\newblock URL \url{https://doi.org/10.1007/s11219-021-09564-z}.

\bibitem[Amit et~al.(2017)Amit, Firstenberg, and Meshi]{archimedes}
Idan Amit, Eyal Firstenberg, and Yinnon Meshi.
\newblock Framework for semi-supervised learning when no labeled data is given.
\newblock U.S. patent application \#US20190164086A1, 2017.
\newblock URL \url{https://patents.google.com/patent/US20190164086A1/}.

\bibitem[Amit et~al.(2019)Amit, Matherly, Hewlett, Xu, Meshi, and
  Weinberger]{amit2019machine}
Idan Amit, John Matherly, William Hewlett, Zhi Xu, Yinnon Meshi, and Yigal
  Weinberger.
\newblock Machine learning in cyber-security - problems, challenges and data
  sets.
\newblock \emph{arXiv preprint arXiv:1812.07858}, 2019.

\bibitem[Amit et~al.(2021)Amit, Ezra, and Feitelson]{amit2021follow}
Idan Amit, Nili~Ben Ezra, and Dror~G. Feitelson.
\newblock Follow your nose -- which code smells are worth chasing?
\newblock \emph{arXiv preprint arXiv:2103.01861}, 2021.

\bibitem[Arpit et~al.(2017)Arpit, Jastrz{\c{e}}bski, Ballas, Krueger, Bengio,
  Kanwal, Maharaj, Fischer, Courville, Bengio, et~al.]{arpit2017closer}
Devansh Arpit, Stanis{\l}aw Jastrz{\c{e}}bski, Nicolas Ballas, David Krueger,
  Emmanuel Bengio, Maxinder~S Kanwal, Tegan Maharaj, Asja Fischer, Aaron
  Courville, Yoshua Bengio, et~al.
\newblock A closer look at memorization in deep networks.
\newblock \emph{arXiv preprint arXiv:1706.05394}, 2017.

\bibitem[Bhandari et~al.(2020)Bhandari, Gour, Ashfaq, Liu, and
  Neubig]{bhandari2020evaluating}
Manik Bhandari, Pranav~Narayan Gour, Atabak Ashfaq, Pengfei Liu, and Graham
  Neubig.
\newblock Re-evaluating evaluation in text summarization.
\newblock In \emph{Proceedings of the 2020 Conference on Empirical Methods in
  Natural Language Processing (EMNLP)}, pages 9347--9359, Online, November
  2020. Association for Computational Linguistics.
\newblock \doi{10.18653/v1/2020.emnlp-main.751}.
\newblock URL \url{https://www.aclweb.org/anthology/2020.emnlp-main.751}.

\bibitem[Chen et~al.(2020)Chen, Syed, Stein, Hagen, and Potthast]{chen:2020a}
Wei-Fan Chen, Shahbaz Syed, Benno Stein, Matthias Hagen, and Martin Potthast.
\newblock {Abstractive Snippet Generation}.
\newblock In Yennung Huang, Irwin King, {Tie-Yan} Liu, and Maarten {van Steen},
  editors, \emph{Web Conference (WWW 2020)}, pages 1309--1319. ACM, April 2020.
\newblock ISBN 978-1-4503-7023-3.
\newblock \doi{10.1145/3366423.3380206}.
\newblock URL \url{https://dl.acm.org/doi/abs/10.1145/3366423.3380206}.

\bibitem[Chopra et~al.(2016)Chopra, Auli, and Rush]{chopra2016abstractive}
Sumit Chopra, Michael Auli, and Alexander~M Rush.
\newblock Abstractive sentence summarization with attentive recurrent neural
  networks.
\newblock In \emph{Proceedings of the 2016 Conference of the North American
  Chapter of the Association for Computational Linguistics: Human Language
  Technologies}, pages 93--98, 2016.

\bibitem[Choshen and Abend(2018{\natexlab{a}})]{Choshen2018AutomaticMV}
Leshem Choshen and Omri Abend.
\newblock Automatic metric validation for grammatical error correction.
\newblock In \emph{ACL}, 2018{\natexlab{a}}.

\bibitem[Choshen and Abend(2018{\natexlab{b}})]{choshen2018inherent}
Leshem Choshen and Omri Abend.
\newblock Inherent biases in reference-based evaluation for grammatical error
  correction and text simplification.
\newblock In \emph{Proceedings of the 56th Annual Meeting of the Association
  for Computational Linguistics (Volume 1: Long Papers)}, pages 632--642,
  Melbourne, Australia, July 2018{\natexlab{b}}. Association for Computational
  Linguistics.
\newblock \doi{10.18653/v1/P18-1059}.
\newblock URL \url{https://www.aclweb.org/anthology/P18-1059}.

\bibitem[Choshen and Abend(2018{\natexlab{c}})]{choshen2018reference}
Leshem Choshen and Omri Abend.
\newblock Reference-less measure of faithfulness for grammatical error
  correction.
\newblock In \emph{Proceedings of the 2018 Conference of the North {A}merican
  Chapter of the Association for Computational Linguistics: Human Language
  Technologies, Volume 2 (Short Papers)}, pages 124--129, New Orleans,
  Louisiana, June 2018{\natexlab{c}}. Association for Computational
  Linguistics.
\newblock \doi{10.18653/v1/N18-2020}.
\newblock URL \url{https://www.aclweb.org/anthology/N18-2020}.

\bibitem[Choshen and Abend(2019)]{choshen2019automatically}
Leshem Choshen and Omri Abend.
\newblock Automatically extracting challenge sets for non-local phenomena in
  neural machine translation.
\newblock In \emph{Proceedings of the 23rd Conference on Computational Natural
  Language Learning (CoNLL)}, pages 291--303, Hong Kong, China, November 2019.
  Association for Computational Linguistics.
\newblock \doi{10.18653/v1/K19-1028}.
\newblock URL \url{https://www.aclweb.org/anthology/K19-1028}.

\bibitem[Choshen and Amit(2021{\natexlab{a}})]{Leshem2021ComsumData}
Leshem Choshen and Idan Amit.
\newblock Comsum: Commit messages summarization and meaning preservation data
  set, 2021{\natexlab{a}}.
\newblock URL
  \url{https://figshare.com/articles/dataset/CumSum_data_set/14711370}.

\bibitem[Choshen and Amit(2021{\natexlab{b}})]{Leshem2021ComsumSrc}
Leshem Choshen and Idan Amit.
\newblock Comsum: Commit messages summarization and meaning preservation source
  code.
\newblock Jul 2021{\natexlab{b}}.
\newblock \doi{10.5281/zenodo.5090706}.

\bibitem[Cohan et~al.(2018)Cohan, Dernoncourt, Kim, Bui, Kim, Chang, and
  Goharian]{cohan2018discourse}
Arman Cohan, Franck Dernoncourt, Doo~Soon Kim, Trung Bui, Seokhwan Kim, Walter
  Chang, and Nazli Goharian.
\newblock A discourse-aware attention model for abstractive summarization of
  long documents.
\newblock \emph{arXiv preprint arXiv:1804.05685}, 2018.

\bibitem[Dou et~al.(2021)Dou, Liu, Hayashi, Jiang, and Neubig]{dou2021gsum}
Zi-Yi Dou, Pengfei Liu, Hiroaki Hayashi, Zhengbao Jiang, and Graham Neubig.
\newblock {GS}um: A general framework for guided neural abstractive
  summarization.
\newblock In \emph{Proceedings of the 2021 Conference of the North American
  Chapter of the Association for Computational Linguistics: Human Language
  Technologies}, pages 4830--4842, Online, June 2021. Association for
  Computational Linguistics.
\newblock URL \url{https://www.aclweb.org/anthology/2021.naacl-main.384}.

\bibitem[Feldman(2020)]{feldman2020does}
Vitaly Feldman.
\newblock Does learning require memorization? a short tale about a long tail.
\newblock In \emph{Proceedings of the 52nd Annual ACM SIGACT Symposium on
  Theory of Computing}, pages 954--959, 2020.

\bibitem[Filippova and Altun(2013)]{filippova2013overcoming}
Katja Filippova and Yasemin Altun.
\newblock Overcoming the lack of parallel data in sentence compression.
\newblock In \emph{Proceedings of the 2013 Conference on Empirical Methods in
  Natural Language Processing}, pages 1481--1491, Seattle, Washington, USA,
  October 2013. Association for Computational Linguistics.
\newblock URL \url{https://www.aclweb.org/anthology/D13-1155}.

\bibitem[Gabriel et~al.(2020)Gabriel, Celikyilmaz, Jha, Choi, and
  Gao]{gabriel2020go}
Saadia Gabriel, Asli Celikyilmaz, Rahul Jha, Yejin Choi, and Jianfeng Gao.
\newblock Go figure! a meta evaluation of factuality in summarization.
\newblock \emph{arXiv preprint arXiv:2010.12834}, 2020.

\bibitem[GitHub(2019)]{bigQuery}
GitHub.
\newblock Github activity data, September 2019.
\newblock URL
  \url{https://console.cloud.google.com/marketplace/product/github/github-repos}.

\bibitem[Graham et~al.(2015)Graham, Baldwin, and Mathur]{graham2015accurate}
Yvette Graham, Timothy Baldwin, and Nitika Mathur.
\newblock Accurate evaluation of segment-level machine translation metrics.
\newblock In \emph{Proceedings of the 2015 Conference of the North {A}merican
  Chapter of the Association for Computational Linguistics: Human Language
  Technologies}, pages 1183--1191, Denver, Colorado, May{--}June 2015.
  Association for Computational Linguistics.
\newblock \doi{10.3115/v1/N15-1124}.
\newblock URL \url{https://www.aclweb.org/anthology/N15-1124}.

\bibitem[Grusky et~al.(2018)Grusky, Naaman, and Artzi]{grusky2018newsroom}
Max Grusky, Mor Naaman, and Yoav Artzi.
\newblock {N}ewsroom: A dataset of 1.3 million summaries with diverse
  extractive strategies.
\newblock In \emph{Proceedings of the 2018 Conference of the North {A}merican
  Chapter of the Association for Computational Linguistics: Human Language
  Technologies, Volume 1 (Long Papers)}, pages 708--719, New Orleans,
  Louisiana, June 2018. Association for Computational Linguistics.
\newblock \doi{10.18653/v1/N18-1065}.
\newblock URL \url{https://www.aclweb.org/anthology/N18-1065}.

\bibitem[Gu et~al.(2018)Gu, Zhang, and Kim]{gu2018deep}
Xiaodong Gu, Hongyu Zhang, and Sunghun Kim.
\newblock Deep code search.
\newblock In \emph{2018 IEEE/ACM 40th International Conference on Software
  Engineering (ICSE)}, pages 933--944. IEEE, 2018.

\bibitem[Gupta and Lehal(2010)]{gupta2010survey}
Vishal Gupta and Gurpreet~Singh Lehal.
\newblock A survey of text summarization extractive techniques.
\newblock \emph{Journal of emerging technologies in web intelligence},
  2\penalty0 (3):\penalty0 258--268, 2010.

\bibitem[Hayati et~al.(2018)Hayati, Olivier, Avvaru, Yin, Tomasic, and
  Neubig]{hayati2018retrieval}
Shirley~Anugrah Hayati, Raphael Olivier, Pravalika Avvaru, Pengcheng Yin,
  Anthony Tomasic, and Graham Neubig.
\newblock Retrieval-based neural code generation.
\newblock In \emph{Proceedings of the 2018 Conference on Empirical Methods in
  Natural Language Processing}, pages 925--930, Brussels, Belgium,
  October-November 2018. Association for Computational Linguistics.
\newblock \doi{10.18653/v1/D18-1111}.
\newblock URL \url{https://www.aclweb.org/anthology/D18-1111}.

\bibitem[Herbold et~al.(2020)Herbold, Trautsch, Ledel, Aghamohammadi, Ghaleb,
  Chahal, Bossenmaier, Nagaria, Makedonski, Ahmadabadi, Szabados, Spieker,
  Madeja, Hoy, Lenarduzzi, Wang, Rodríguez-Pérez, Colomo-Palacios,
  Verdecchia, Singh, Qin, Chakroborti, Davis, Walunj, Wu, Marcilio, Alam,
  Aldaeej, Amit, Turhan, Eismann, Wickert, Malavolta, Sulir, Fard, Henley,
  Kourtzanidis, Tuzun, Treude, Shamasbi, Pashchenko, Wyrich, Davis, Serebrenik,
  Albrecht, Aktas, Strüber, and Erbel]{herbold2020largescale}
Steffen Herbold, Alexander Trautsch, Benjamin Ledel, Alireza Aghamohammadi,
  Taher~Ahmed Ghaleb, Kuljit~Kaur Chahal, Tim Bossenmaier, Bhaveet Nagaria,
  Philip Makedonski, Matin~Nili Ahmadabadi, Kristof Szabados, Helge Spieker,
  Matej Madeja, Nathaniel Hoy, Valentina Lenarduzzi, Shangwen Wang, Gema
  Rodríguez-Pérez, Ricardo Colomo-Palacios, Roberto Verdecchia, Paramvir
  Singh, Yihao Qin, Debasish Chakroborti, Willard Davis, Vijay Walunj, Hongjun
  Wu, Diego Marcilio, Omar Alam, Abdullah Aldaeej, Idan Amit, Burak Turhan,
  Simon Eismann, Anna-Katharina Wickert, Ivano Malavolta, Matus Sulir, Fatemeh
  Fard, Austin~Z. Henley, Stratos Kourtzanidis, Eray Tuzun, Christoph Treude,
  Simin~Maleki Shamasbi, Ivan Pashchenko, Marvin Wyrich, James Davis, Alexander
  Serebrenik, Ella Albrecht, Ethem~Utku Aktas, Daniel Strüber, and Johannes
  Erbel.
\newblock Large-scale manual validation of bug fixing commits: A fine-grained
  analysis of tangling.
\newblock \emph{arXiv preprint arXiv:2011.06244}, 2020.

\bibitem[Hermann et~al.(2015)Hermann, Kocisky, Grefenstette, Espeholt, Kay,
  Suleyman, and Blunsom]{48562}
Karl~Moritz Hermann, Tomas Kocisky, Edward Grefenstette, Lasse Espeholt, Will
  Kay, Mustafa Suleyman, and Phil Blunsom.
\newblock Teaching machines to read and comprehend.
\newblock In \emph{NIPS}, 2015.

\bibitem[{Herzig} and {Zeller}(2013)]{6624018}
K.~{Herzig} and A.~{Zeller}.
\newblock The impact of tangled code changes.
\newblock In \emph{2013 10th Working Conference on Mining Software Repositories
  (MSR)}, pages 121--130, 2013.
\newblock \doi{10.1109/MSR.2013.6624018}.

\bibitem[Honovich et~al.(2021)Honovich, Choshen, Aharoni, Neeman, Szpektor, and
  Abend]{honovich2021q}
Or~Honovich, Leshem Choshen, Roee Aharoni, Ella Neeman, Idan Szpektor, and Omri
  Abend.
\newblock {$\mathbf{Q}^2$}: Evaluating factual consistency in
  knowledge-grounded dialogues via question generation and question answering.
\newblock \emph{arXiv preprint arXiv:2104.08202}, 2021.

\bibitem[Kryscinski et~al.(2020)Kryscinski, McCann, Xiong, and
  Socher]{kryscinski2020evaluating}
Wojciech Kryscinski, Bryan McCann, Caiming Xiong, and Richard Socher.
\newblock Evaluating the factual consistency of abstractive text summarization.
\newblock In \emph{Proceedings of the 2020 Conference on Empirical Methods in
  Natural Language Processing (EMNLP)}, pages 9332--9346, Online, November
  2020. Association for Computational Linguistics.
\newblock \doi{10.18653/v1/2020.emnlp-main.750}.
\newblock URL \url{https://www.aclweb.org/anthology/2020.emnlp-main.750}.

\bibitem[Kry{\'s}ci{\'n}ski et~al.(2021)Kry{\'s}ci{\'n}ski, Rajani, Agarwal,
  Xiong, and Radev]{kryscinski2021booksum}
Wojciech Kry{\'s}ci{\'n}ski, Nazneen Rajani, Divyansh Agarwal, Caiming Xiong,
  and Dragomir Radev.
\newblock Booksum: A collection of datasets for long-form narrative
  summarization.
\newblock \emph{arXiv preprint arXiv:2105.08209}, 2021.

\bibitem[Lachmy et~al.(2021)Lachmy, Yao, Durrett, Gligoric, Li, Mooney, Neubig,
  Su, Sun, and Tsarfaty]{nlp4prog}
Royi Lachmy, Ziyu Yao, Greg Durrett, Milos Gligoric, Junyi~Jessy Li, Ray
  Mooney, Graham Neubig, Yu~Su, Huan Sun, and Reut Tsarfaty.
\newblock The first workshop on natural language processing for programming,
  August 2021.
\newblock URL \url{https://nlp4prog.github.io/2021/cfp/}.

\bibitem[Lewis et~al.(2020)Lewis, Liu, Goyal, Ghazvininejad, Mohamed, Levy,
  Stoyanov, and Zettlemoyer]{lewis2020bart}
Mike Lewis, Yinhan Liu, Naman Goyal, Marjan Ghazvininejad, Abdelrahman Mohamed,
  Omer Levy, Veselin Stoyanov, and Luke Zettlemoyer.
\newblock {BART}: Denoising sequence-to-sequence pre-training for natural
  language generation, translation, and comprehension.
\newblock In \emph{Proceedings of the 58th Annual Meeting of the Association
  for Computational Linguistics}, pages 7871--7880, Online, July 2020.
  Association for Computational Linguistics.
\newblock \doi{10.18653/v1/2020.acl-main.703}.
\newblock URL \url{https://www.aclweb.org/anthology/2020.acl-main.703}.

\bibitem[Lin(2004)]{lin2004rouge}
Chin-Yew Lin.
\newblock {ROUGE}: A package for automatic evaluation of summaries.
\newblock In \emph{Text Summarization Branches Out}, pages 74--81, Barcelona,
  Spain, July 2004. Association for Computational Linguistics.
\newblock URL \url{https://www.aclweb.org/anthology/W04-1013}.

\bibitem[Macketanz et~al.(2018)Macketanz, Avramidis, Burchardt, and
  Uszkoreit]{macketanz2018fine}
Vivien Macketanz, Eleftherios Avramidis, Aljoscha Burchardt, and Hans
  Uszkoreit.
\newblock Fine-grained evaluation of {G}erman-{E}nglish machine translation
  based on a test suite.
\newblock In \emph{Proceedings of the Third Conference on Machine Translation:
  Shared Task Papers}, pages 578--587, Belgium, Brussels, October 2018.
  Association for Computational Linguistics.
\newblock \doi{10.18653/v1/W18-6436}.
\newblock URL \url{https://www.aclweb.org/anthology/W18-6436}.

\bibitem[Miceli~Barone and Sennrich(2017)]{miceli2017parallel}
Antonio~Valerio Miceli~Barone and Rico Sennrich.
\newblock A parallel corpus of python functions and documentation strings for
  automated code documentation and code generation.
\newblock In \emph{Proceedings of the Eighth International Joint Conference on
  Natural Language Processing (Volume 2: Short Papers)}, pages 314--319,
  Taipei, Taiwan, November 2017. Asian Federation of Natural Language
  Processing.
\newblock URL \url{https://www.aclweb.org/anthology/I17-2053}.

\bibitem[Mockus and Votta(2000)]{mockus2000identifying}
Audris Mockus and Lawrence~G Votta.
\newblock Identifying reasons for software changes using historic databases.
\newblock In \emph{icsm}, pages 120--130, 2000.

\bibitem[Nallapati et~al.(2016)Nallapati, Zhou, Santos, Çaglar G{\"u}lçehre,
  and Xiang]{Nallapati2016AbstractiveTS}
Ramesh Nallapati, Bowen Zhou, C.~D. Santos, Çaglar G{\"u}lçehre, and Bing
  Xiang.
\newblock Abstractive text summarization using sequence-to-sequence rnns and
  beyond.
\newblock In \emph{CoNLL}, 2016.

\bibitem[Narayan et~al.(2018)Narayan, Cohen, and Lapata]{narayan2018dont}
Shashi Narayan, Shay~B. Cohen, and Mirella Lapata.
\newblock Don{'}t give me the details, just the summary! topic-aware
  convolutional neural networks for extreme summarization.
\newblock In \emph{Proceedings of the 2018 Conference on Empirical Methods in
  Natural Language Processing}, pages 1797--1807, Brussels, Belgium,
  October-November 2018. Association for Computational Linguistics.
\newblock \doi{10.18653/v1/D18-1206}.
\newblock URL \url{https://www.aclweb.org/anthology/D18-1206}.

\bibitem[Oak et~al.(2019)Oak, Du, Yan, Takawale, and Amit]{oak2019malware}
Rajvardhan Oak, Min Du, David Yan, Harshvardhan Takawale, and Idan Amit.
\newblock Malware detection on highly imbalanced data through sequence
  modeling.
\newblock In \emph{Proceedings of the 12th ACM Workshop on Artificial
  Intelligence and Security}, pages 37--48, 2019.

\bibitem[Over et~al.(2007)Over, Dang, and Harman]{over2007duc}
Paul Over, Hoa Dang, and Donna Harman.
\newblock Duc in context.
\newblock \emph{Information Processing \& Management}, 43\penalty0
  (6):\penalty0 1506--1520, 2007.

\bibitem[Pagnoni et~al.(2021)Pagnoni, Balachandran, and
  Tsvetkov]{pagnoni2021understanding}
Artidoro Pagnoni, Vidhisha Balachandran, and Yulia Tsvetkov.
\newblock Understanding factuality in abstractive summarization with {FRANK}: A
  benchmark for factuality metrics.
\newblock In \emph{Proceedings of the 2021 Conference of the North American
  Chapter of the Association for Computational Linguistics: Human Language
  Technologies}, pages 4812--4829, Online, June 2021. Association for
  Computational Linguistics.
\newblock URL \url{https://www.aclweb.org/anthology/2021.naacl-main.383}.

\bibitem[Panthaplackel et~al.(2020)Panthaplackel, Nie, Gligoric, Li, and
  Mooney]{panthaplackel2020learning}
Sheena Panthaplackel, Pengyu Nie, Milos Gligoric, Junyi~Jessy Li, and Raymond
  Mooney.
\newblock Learning to update natural language comments based on code changes.
\newblock In \emph{Proceedings of the 58th Annual Meeting of the Association
  for Computational Linguistics}, pages 1853--1868, Online, July 2020.
  Association for Computational Linguistics.
\newblock \doi{10.18653/v1/2020.acl-main.168}.
\newblock URL \url{https://www.aclweb.org/anthology/2020.acl-main.168}.

\bibitem[Qi et~al.(2020)Qi, Yan, Gong, Liu, Duan, Chen, Zhang, and
  Zhou]{qi2020prophetnet}
Weizhen Qi, Yu~Yan, Yeyun Gong, Dayiheng Liu, Nan Duan, Jiusheng Chen, Ruofei
  Zhang, and Ming Zhou.
\newblock {P}rophet{N}et: Predicting future n-gram for
  sequence-to-{S}equence{P}re-training.
\newblock In \emph{Findings of the Association for Computational Linguistics:
  EMNLP 2020}, pages 2401--2410, Online, November 2020. Association for
  Computational Linguistics.
\newblock \doi{10.18653/v1/2020.findings-emnlp.217}.
\newblock URL \url{https://www.aclweb.org/anthology/2020.findings-emnlp.217}.

\bibitem[Ratner et~al.(2016)Ratner, De~Sa, Wu, Selsam, and
  R\'{e}]{NIPS2016_6523}
Alexander~J Ratner, Christopher~M De~Sa, Sen Wu, Daniel Selsam, and Christopher
  R\'{e}.
\newblock Data programming: Creating large training sets, quickly.
\newblock In D.~D. Lee, M.~Sugiyama, U.~V. Luxburg, I.~Guyon, and R.~Garnett,
  editors, \emph{Advances in Neural Information Processing Systems 29}, pages
  3567--3575. Curran Associates, Inc., 2016.
\newblock URL
  \url{http://papers.nips.cc/paper/6523-data-programming-creating-large-training-sets-quickly.pdf}.

\bibitem[Ribeiro et~al.(2020)Ribeiro, Wu, Guestrin, and
  Singh]{ribeiro2020beyond}
Marco~Tulio Ribeiro, Tongshuang Wu, Carlos Guestrin, and Sameer Singh.
\newblock Beyond accuracy: Behavioral testing of {NLP} models with
  {C}heck{L}ist.
\newblock In \emph{Proceedings of the 58th Annual Meeting of the Association
  for Computational Linguistics}, pages 4902--4912, Online, July 2020.
  Association for Computational Linguistics.
\newblock \doi{10.18653/v1/2020.acl-main.442}.
\newblock URL \url{https://www.aclweb.org/anthology/2020.acl-main.442}.

\bibitem[Rush et~al.(2015)Rush, Chopra, and Weston]{rush2015neural}
Alexander~M. Rush, Sumit Chopra, and Jason Weston.
\newblock A neural attention model for abstractive sentence summarization.
\newblock In \emph{Proceedings of the 2015 Conference on Empirical Methods in
  Natural Language Processing}, pages 379--389, Lisbon, Portugal, September
  2015. Association for Computational Linguistics.
\newblock \doi{10.18653/v1/D15-1044}.
\newblock URL \url{https://www.aclweb.org/anthology/D15-1044}.

\bibitem[Sandhaus(2008)]{sandhaus2008new}
Evan Sandhaus.
\newblock The new york times annotated corpus.
\newblock \emph{Linguistic Data Consortium, Philadelphia}, 6\penalty0
  (12):\penalty0 e26752, 2008.

\bibitem[Santos and Hindle(2016)]{santos2016judging}
Eddie~Antonio Santos and Abram Hindle.
\newblock Judging a commit by its cover.
\newblock In \emph{Proceedings of the 13th International Workshop on Mining
  Software Repositories-MSR}, volume~16, pages 504--507, 2016.

\bibitem[Schapire(1990)]{schapire1990strength}
Robert~E Schapire.
\newblock The strength of weak learnability.
\newblock \emph{Machine learning}, 5\penalty0 (2):\penalty0 197--227, 1990.

\bibitem[Settles(2010)]{Settles10activelearning}
Burr Settles.
\newblock Active learning literature survey.
\newblock Technical report, University of Wisconsin–Madison, 2010.

\bibitem[Sharma et~al.(2019)Sharma, Li, and Wang]{sharma2019bigpatent}
Eva Sharma, Chen Li, and Lu~Wang.
\newblock Bigpatent: A large-scale dataset for abstractive and coherent
  summarization.
\newblock \emph{arXiv preprint arXiv:1906.03741}, 2019.

\bibitem[Slobodkin et~al.(2021)Slobodkin, Choshen, and
  Abend]{slobodkin2021mediators}
Aviv Slobodkin, Leshem Choshen, and Omri Abend.
\newblock Mediators in determining what processing {BERT} performs first.
\newblock In \emph{Proceedings of the 2021 Conference of the North American
  Chapter of the Association for Computational Linguistics: Human Language
  Technologies}, pages 86--93, Online, June 2021. Association for Computational
  Linguistics.
\newblock URL \url{https://www.aclweb.org/anthology/2021.naacl-main.8}.

\bibitem[Swanson(1976)]{swanson1976dimensions}
E~Burton Swanson.
\newblock The dimensions of maintenance.
\newblock In \emph{Proceedings of the 2nd international conference on Software
  engineering}, pages 492--497, 1976.

\bibitem[Tenney et~al.(2019)Tenney, Das, and Pavlick]{tenney2019bert}
Ian Tenney, Dipanjan Das, and Ellie Pavlick.
\newblock {BERT} rediscovers the classical {NLP} pipeline.
\newblock In \emph{Proceedings of the 57th Annual Meeting of the Association
  for Computational Linguistics}, pages 4593--4601, Florence, Italy, July 2019.
  Association for Computational Linguistics.
\newblock \doi{10.18653/v1/P19-1452}.
\newblock URL \url{https://www.aclweb.org/anthology/P19-1452}.

\bibitem[Terrell et~al.(2017)Terrell, Kofink, Middleton, Rainear, Murphy-Hill,
  Parnin, and Stallings]{terrell2017gender}
Josh Terrell, Andrew Kofink, Justin Middleton, Clarissa Rainear, Emerson
  Murphy-Hill, Chris Parnin, and Jon Stallings.
\newblock Gender differences and bias in open source: Pull request acceptance
  of women versus men.
\newblock \emph{PeerJ Computer Science}, 3:\penalty0 e111, 2017.

\bibitem[Vadlapudi and Katragadda(2010)]{vadlapudi2010automated}
Ravikiran Vadlapudi and Rahul Katragadda.
\newblock On automated evaluation of readability of summaries: Capturing
  grammaticality, focus, structure and coherence.
\newblock In \emph{Proceedings of the {NAACL} {HLT} 2010 Student Research
  Workshop}, pages 7--12, Los Angeles, CA, June 2010. Association for
  Computational Linguistics.
\newblock URL \url{https://www.aclweb.org/anthology/N10-3002}.

\bibitem[Voita et~al.(2020)Voita, Sennrich, and Titov]{voita2020analyzing}
Elena Voita, Rico Sennrich, and Ivan Titov.
\newblock Analyzing the source and target contributions to predictions in
  neural machine translation.
\newblock \emph{arXiv preprint arXiv:2010.10907}, 2020.

\bibitem[V{\"o}lske et~al.(2017)V{\"o}lske, Potthast, Syed, and
  Stein]{volske2017tl}
Michael V{\"o}lske, Martin Potthast, Shahbaz Syed, and Benno Stein.
\newblock {TL};{DR}: Mining {R}eddit to learn automatic summarization.
\newblock In \emph{Proceedings of the Workshop on New Frontiers in
  Summarization}, pages 59--63, Copenhagen, Denmark, September 2017.
  Association for Computational Linguistics.
\newblock \doi{10.18653/v1/W17-4508}.
\newblock URL \url{https://www.aclweb.org/anthology/W17-4508}.

\bibitem[Warner(2018)]{warner100mrepo}
Jason Warner.
\newblock Thank you for 100 million repositories.
\newblock \emph{The GitHub Blog}, November 2018.
\newblock URL \url{https://github.blog/2018-11-08-100m-repos/}.

\bibitem[Warstadt et~al.(2020)Warstadt, Parrish, Liu, Mohananey, Peng, Wang,
  and Bowman]{warstadt2020blimp}
Alex Warstadt, Alicia Parrish, Haokun Liu, Anhad Mohananey, Wei Peng, Sheng-Fu
  Wang, and Samuel~R. Bowman.
\newblock {BL}i{MP}: The benchmark of linguistic minimal pairs for {E}nglish.
\newblock \emph{Transactions of the Association for Computational Linguistics},
  8:\penalty0 377--392, 2020.
\newblock \doi{10.1162/tacl_a_00321}.
\newblock URL \url{https://www.aclweb.org/anthology/2020.tacl-1.25}.

\bibitem[Wilber et~al.(2021)Wilber, Timkey, and van Schijndel]{Wilber2021ToPO}
Matthew Wilber, William Timkey, and Marten van Schijndel.
\newblock To point or not to point: Understanding how abstractive summarizers
  paraphrase text.
\newblock 2021.

\bibitem[Xu and Durrett(2021)]{Xu2021DissectingGM}
Jiacheng Xu and Greg Durrett.
\newblock Dissecting generation modes for abstractive summarization models via
  ablation and attribution.
\newblock 2021.

\bibitem[Zhang et~al.(2020)Zhang, Zhao, Saleh, and Liu]{zhang2020pegasus}
Jingqing Zhang, Yao Zhao, Mohammad Saleh, and Peter Liu.
\newblock Pegasus: Pre-training with extracted gap-sentences for abstractive
  summarization.
\newblock In \emph{International Conference on Machine Learning}, pages
  11328--11339. PMLR, 2020.

\bibitem[Zhang et~al.(2019)Zhang, Kishore, Wu, Weinberger, and
  Artzi]{zhang2019bertscore}
Tianyi Zhang, Varsha Kishore, Felix Wu, Kilian~Q Weinberger, and Yoav Artzi.
\newblock Bertscore: Evaluating text generation with bert.
\newblock In \emph{International Conference on Learning Representations}, 2019.

\end{thebibliography}

\FloatBarrier
\appendix

\section{Supplementary Materials}\label{sec:sup}

Source code and documentation are available at \cite{Leshem2021ComsumSrc} and \url{https://github.com/evidencebp/comsum}.
Data is available at \cite{Leshem2021ComsumData}.
Analysis used the infrastructure for GitHub \cite{Amit2021General}, natural language processing for software engineering  \cite{Amit2021CommitClassification}, and analysis utilities \cite{Amit2021Analysis}.

\section{Summary} \label{ap:git_sum}
\begin{figure}[tbp]
\includegraphics[width=8cm]{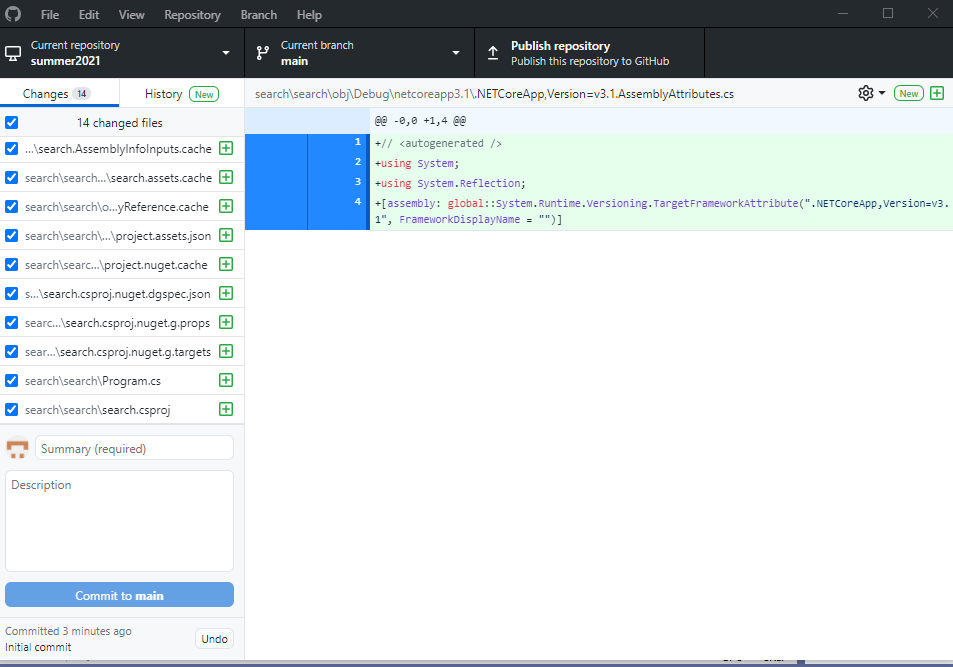}
\caption{Git's Graphical User Interface where they leave a place for the developer to fill a message and a summary.}
\label{fig:git_sum}
\end{figure}

Git describes the one line as a summary, or sometimes as a subject or header (similar to news where titles are considered summarization). We bring the git Graphical User interface as an example of it in Fig.~\ref{fig:git_sum}, we bring further examples in
\url{https://github.com/evidencebp/comsum/tree/main/data/images}.

\section{Labeling protocol}\label{ap:guidelines}
Summary labeling protocol

We label for two concept - Is\_summary and Is\_Generic Specific cases

\begin{enumerate}
\def\labelenumi{\arabic{enumi}.}
\item
  Summary

  \begin{enumerate}
  \def\labelenumii{\arabic{enumii}.}
  \itemsep1pt\parskip0pt\parsep0pt
  \item
    We consider only English messages (99\% of the commits). The messgae
    must be in English to be considered a summary.
  \item
    A summary should reflect information in the message, \textbf{not} to
    add \textbf{new information}.
  \item
    In a summary, the subject should capture the \textbf{essence} of the
    message.
  \item
    Specific cases:

    \begin{enumerate}
    \def\labelenumiii{\arabic{enumiii}.}
    \itemsep1pt\parskip0pt\parsep0pt
    \item
      If the subject appears in the rest of the message (as in
      extractive summarization), this is a summary.
    \item
      Change/detail, the subject describes a change and the message
      describes its details, it is a summary.
    \item
      If the commit classification of the subject is different from the
      message, this is not a summary.
    \item
      Goal/mean relation (e.g. ``prevent crash'' \& ``check amount is
      not zero before computing average price''), is considered as a
      summary.
    \item
      Change/reason (e.g., ``added caller information'' \& ``..to ease
      debug'') is considered to be a summary.
    \item
      Administrative message, if all the changed content is in the
      subject (e.g., the message contains only administrative
      information like ticket number and annotator name) , is not a
      summary.
    \item
      Merge subject (``Merge branch `15.3'\,'') and a message that
      describes the content is not a summary. That is since the content
      could fit branch `16.3' just as well.
    \item
      Content announcing subject (``Declaration of'') and a message that
      describes the content is not a summary. That is since different
      contents could fit `Declaration of' just as well.
    \item
      Subject content by reference (e.g., just referring to a ticket
      number) is not a summary. While the content exists, it is not self
      contained.
    \item
      If the message content is only by reference (e.g., hash of the
      reverted commit), it is not self-contained and cannot be
      summarized. Hene, this is not a summary.
    \end{enumerate}
  \end{enumerate}
\item
  Generic

  \begin{enumerate}
  \def\labelenumii{\arabic{enumii}.}
  \itemsep1pt\parskip0pt\parsep0pt
  \item
    If the subject can be applied to many other commits it is generic.
  \item
    Specific cases:

    \begin{enumerate}
    \def\labelenumiii{\arabic{enumiii}.}
    \itemsep1pt\parskip0pt\parsep0pt
    \item
      Subject: ``WIP'' (work in progress) is generic and not a summary.
    \item
      If the subject can be applied to many other commits, yet fully
      describe the change (e.g., `Updated Spanish translation'), it is
      generic and summary.
    \item
      Merge subject (``Merge branch `15.3'\,'') is generic.
    \item
      Content announcing subject (``Declaration of'') is generic.
    \end{enumerate}
  \end{enumerate}
\end{enumerate}

\section{Labeling for the administrative heuristic}\label{ap:ad_heur}

Commit message can be viewed as containing content describing the code change (e.g., `extracted method') and administrative content (e.g., 'Signed-off-by: Alan Turing').

The administrative content usually uses few specific terms that can be identified.
Our heuristic looks for these terms in the message and classify it as administrative if an administrative term appears in the first 20 characters.
The intuition of the heuristic is that 20 characters do not leave space for code change description.
We labeled all 265 hits in 5,000 commits samples.
2 samples were summaries with a change/details relation.
1 sample was a merge.
98.9\% of the matches needed removing.
46 samples, with distance closer to 20 were reference commits (e.g., fixed bug \#123).
These are also not suitable for summary and should be removed.
The hit rate is 5.3\% compared to a 7\% positive rate in the random sample, indicating a recall of about 75\%.
All administrative commits in the random samples were identified.
\section{BART hyperparameters}\label{ap:hyperparams}
For finetuning BART we used max target length of 128 and source length of 512, learning rate of $1e^{-4}$ and 256 batch size. Other parameters including the zero-shot's are the defaults by the \href{https://huggingface.co/}{HuggingFace} library. The model was trained for a week on 4 Nvidia M60 GPUs.

\section{Ethical Considerations}\label{sec:ethical}

The messages contained in the data set were written by \authorsNum developers contributing to open source projects.
We could not get their \emph{direct} approval to use the messages in the data set.
However, open-source projects allow not only access to the commit messages but even to the source code.
Developers are aware of that and agree to it as it is a part of the development project of all \emph{public open source} projects.
We validated in GitHub and all the projects included in ComSum have an OSI-approved open source license.

While we do not store developers' personal information, each commit is identified by a hash.
Given the hash, a lookup in the project metadata retrieves the developer's profile.
Since it is required from the development process, the developers accept that and we do not ease lookup or provide new information about the developer. In any case, the developer controls the data published on them and not us. Moreover, they can remove or alter it in any way that does not violate GitHub's terms. We consider this concern as addressed too.

7K commits were identified using classifiers from \cite{Amit2021CCP}, to contain swearing and 325k commits were identified to contain negative sentiment.
The true numbers might be higher due to the classifiers' false negatives. As this data is already open we did not filter those, but warn users to filter profanity if their needs so require.

\section{Hosting, Licensing, and Maintenance plan}

We release the data with the license Creative Commons version 4.0 (aka, CC-4), allowing copy, redistribution and other common uses without the need of permission but with proper credit.
We do not plan to change it.

We host the commit data set in figshare.

As for the maintenance plan, we provide all the code used to generate the data set.

The code will enable any researcher to maintain the data set and keep extending it.
This includes adding data from future work in the projects, using different selection conditions, enhancement with more features, etc.

\end{document}